\documentclass{article} 
\usepackage{iclr2027_conference,times}

\usepackage{amsmath,amsfonts,bm}

\def\eqref#1{equation~\ref{#1}}

\def\1{\bm{1}}

\DeclareMathAlphabet{\mathsfit}{\encodingdefault}{\sfdefault}{m}{sl}
\SetMathAlphabet{\mathsfit}{bold}{\encodingdefault}{\sfdefault}{bx}{n}

\usepackage{hyperref}
\usepackage{url}

\usepackage{graphicx}
\usepackage{natbib}
\usepackage{pifont}
\usepackage{xcolor}
\usepackage{colortbl}
\usepackage{amsmath}
\usepackage{bm}

\usepackage{makecell}
\usepackage{multirow}
\usepackage{tabularx}

\usepackage{amsfonts}

\usepackage{algorithm}
\usepackage{algorithmic}

\usepackage{booktabs}

\usepackage{array}

\newcommand{\cmark}{\textcolor{green!50!black}{\ding{51}}}
\newcommand{\xmark}{\textcolor{red!70!black}{\ding{55}}}

\title{CrossModalQA: A Cross-modal and Multi-hop Benchmark for Multimodal Retrieval-augmented Generation}

\author{Jiacheng Cai$^{1}$, Zijin Hong$^{1}$, Zheng Yuan$^{1}$, 
  Huachi Zhou$^1$, Qinggang Zhang$^2$, Xiao Huang$^{1\dagger}$\\
  $^1$The Hong Kong Polytechnic University,
  $^2$Jilin University \\
  \texttt{\{jiacheng.cai, zijin.hong, yzheng.yuan, huachi.zhou\}}\\
  \texttt{@connect.polyu.hk, }
  \texttt{qinggangzhang@jlu.edu.cn,}\\  \texttt{xiao.huang@polyu.edu.hk} 
}

\usepackage{booktabs}
\usepackage{makecell}
\usepackage{multirow}
\usepackage{tabularx}
\usepackage{array}

\newcolumntype{L}[1]{%
  >{\raggedright\arraybackslash}m{#1}}
\newcolumntype{C}[1]{%
  >{\centering\arraybackslash}m{#1}}

\iclrfinalcopy 
\begin{document}

\maketitle

\begin{abstract}
Despite the strong capabilities of multimodal large language models (MLLMs), their parametric knowledge remains incomplete and difficult to update, motivating multimodal retrieval-augmented generation (RAG) to ground responses in external text and images. 
However, existing benchmarks face two major limitations: (i) they typically emphasize single-hop retrieval or reasoning over a small set of provided contexts rather than open-domain evidence discovery; and (ii) they provide fragmented coverage of cross-modal reasoning paths, leaving complex multi-hop and multi-image reasoning underexplored. 
In this paper, we introduce \textbf{CrossModalQA}, an open-domain benchmark for evaluating multimodal retrieval and reasoning over heterogeneous corpora. 
CrossModalQA contains 1,863 question-answer pairs constructed from 4,987 Wikipedia articles and 4,431 Wikimedia Commons images. 
It covers five complementary reasoning paths: vision-to-text, text-to-vision, vision-to-text-to-vision, multi-image intersection, and image-set reasoning. Every question requires retrieving and composing distributed textual and visual evidence, with an average reasoning depth of 3.50 hops. We construct the benchmark through multimodal knowledge graph-guided subgraph sampling and apply rule-based consistency checking and LLM verification to ensure multimodal dependence and traceable evidence. Extensive experiments demonstrate that existing multimodal RAG systems struggle to recover complete evidence chains and can underperform closed-book models when incomplete retrieval introduces distracting context. Further analysis reveals that complete cross-modal retrieval contributes more to answer accuracy than generator scaling, while multi-image retrieval and reasoning remain the primary bottlenecks limiting end-to-end performance.
\end{abstract}

\begingroup
\renewcommand{\thefootnote}{\ensuremath{\dagger}}
\footnotetext{Corresponding author.}
\endgroup

\section{Introduction}

\begin{figure}[ht]
\centering
\includegraphics[width=0.98\textwidth]{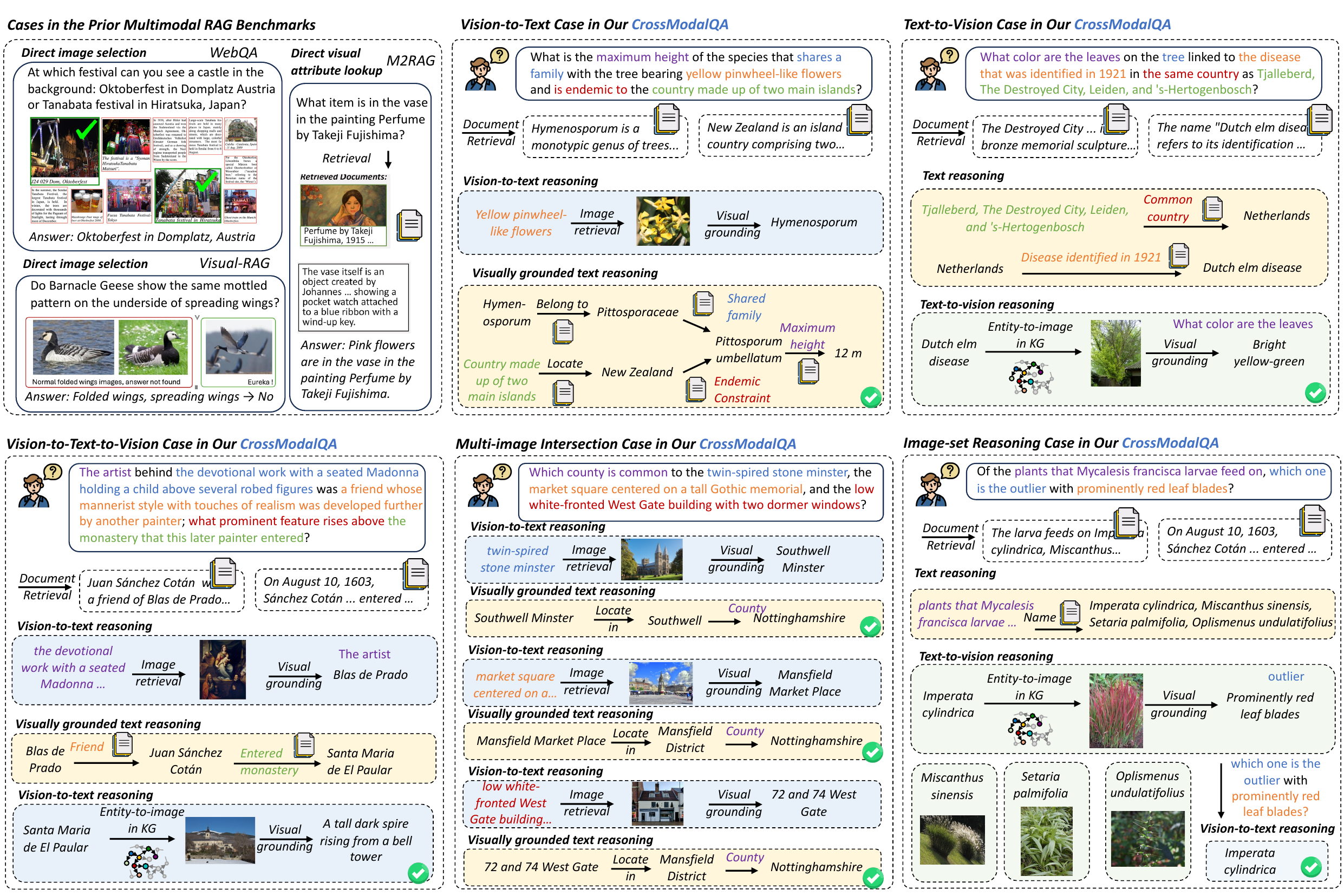} 
\caption{Overview of CrossModalQA. The figure shows examples of five question modes defined in the CrossModalQA. Questions are in the open-domain. We present the input, ground truth answers, and retrieved documents for each task. Compared with other benchmarks, CrossModalQA offers a more fine-grained, modality-aware categorization of retrieval–reasoning paths and requires a larger number of reasoning hops.}
\label{fig:example}
\end{figure}

Large language models (LLMs) have demonstrated strong capabilities in language understanding, knowledge-intensive question answering, and complex reasoning~\citep{brown2020language,achiam2023gpt,touvron2023llama}.
Nevertheless, the knowledge encoded in model parameters is inherently incomplete and difficult to update, which can lead LLMs to generate plausible but factually unsupported responses.
Retrieval-augmented generation (RAG) addresses this limitation by retrieving relevant evidence from an external corpus and conditioning generation on the retrieved context~\citep{lewis2020retrieval,karpukhin2020dense,izacard2021leveraging}.
This retrieve-then-generate paradigm provides access to updatable knowledge and reliably grounds model predictions in external evidence.

Most early RAG systems assume that both queries and external knowledge are represented as text.
This assumption overlooks a substantial portion of real-world knowledge conveyed visually.
Images directly encode appearance, color, shape, symbols, spatial configurations, and other attributes that may be absent from textual documents.
Meanwhile, multimodal large language models (MLLMs) have substantially improved the ability to perceive and reason over visual inputs~\citep{alayrac2022flamingo,li2023blip2,liu2023visual}.
These advances have motivated multimodal RAG, which retrieves textual and visual evidence and supplies it to an MLLM for knowledge-grounded generation~\citep{chen2022murag,chen2024mllm,bai2025ramqa}.
Compared with text-only RAG, multimodal RAG must retrieve relevant evidence from heterogeneous sources, ground entities across modalities, and compose complementary information into a complete reasoning chain.

Existing multimodal QA and RAG benchmarks evaluate different aspects of reasoning over heterogeneous sources.
MultiModalQA and WebQA study reasoning across text, images, tables, or captions~\citep{talmor2021multimodalqa,chang2022webqa}, while RetVQA and Visual-RAG emphasize visual evidence retrieval~\citep{penamakuri2023retvqa,wu2025visual}.
InfoSeek and Encyclopedic-VQA focus on knowledge-intensive questions involving visual and textual information~\citep{chen2023can,mensink2023encyclopedic}.
More recent benchmarks extend evaluation to visually augmented questions, dynamic retrieval, multimodal contexts, and multimodal answers~\citep{hu2025mrag,li2025benchmarking,liu2025benchmarking,yu2025mramg}.

Despite this progress, existing benchmarks face two major limitations, as summarized in Table~\ref{tab:five_modes}.
\textbf{(i) Fragmented reasoning-path coverage}: most benchmarks focus on one or two modality transitions, such as \textit{V-to-T} or \textit{T-to-V}.
More complex structures, including \textit{V-to-T-to-V}, multi-image intersection, and image-set reasoning, remain largely underexplored.
This fragmented coverage makes it difficult to determine whether a system can adapt its retrieval and reasoning behavior to different sequences of textual and visual evidence.
\textbf{(ii) Insufficient evaluation of open-domain multi-hop reasoning}: existing questions are often dominated by short evidence chains or provide a small set of candidate contexts directly.
These settings primarily evaluate local evidence selection or context utilization rather than whether a system can discover and connect multiple evidence items from a large heterogeneous corpus.
A model may therefore succeed through a single modality, a partial retrieval result, or a fixed retrieval pipeline without recovering the complete evidence chain.

Effective multimodal RAG evaluation poses a central challenge: \textbf{measuring whether a system can discover, connect, and compose distributed textual and visual evidence across diverse reasoning paths}.
Addressing this challenge requires a benchmark that jointly controls modality transitions and reasoning depth while preserving realistic open-domain retrieval.
It must also retain explicit supporting evidence so that retrieval failures can be distinguished from reasoning and generation errors.

With this motivation, we introduce \textbf{CrossModalQA}, an open-domain multimodal retrieval-augmented question-answering benchmark containing 1,863 text-only question--answer pairs constructed from 4,987 Wikipedia articles and 4,431 Wikimedia Commons images.
Each question requires retrieving and composing evidence from multiple nodes in a multimodal knowledge graph, where entities and images are connected through textual relations and image--entity links.
Based on the modality sequence and evidence structure of the supporting subgraph, we organize the questions into five reasoning-path categories: \textit{V-to-T}, \textit{T-to-V}, \textit{V-to-T-to-V}, \textit{multi-image intersection}, and \textit{image-set reasoning}.
Here, \textit{V} and \textit{T} denote the modalities traversed during reasoning rather than the input modality.
Although every question is presented as text, its solution requires retrieving and reasoning over visual evidence.

We construct CrossModalQA through domain-targeted article selection, knowledge graph extraction, multimodal grounding, and subgraph-conditioned question generation.
Each question retains an explicit multimodal supporting subgraph that enables its reasoning complexity to be characterized by graph paths and hop counts.
The resulting questions require an average of 3.50 reasoning hops, with more than 40\% requiring at least four hops.
Rule-based consistency checking and LLM verification further ensure that each example follows its intended reasoning path and depends on multimodal evidence.
This construction supports fine-grained evaluation of retrieval, cross-modal evidence linking, reasoning, and answer generation.

Extensive experiments with representative multimodal RAG systems reveal substantial limitations in recovering complete evidence chains.
Existing systems can underperform a closed-book MLLM when incomplete retrieval omits necessary evidence and introduces distracting context.
Oracle experiments show that providing complete textual and visual evidence substantially improves answer accuracy, indicating that accurate retrieval contributes more than simply scaling the answer generator.
Further analysis reveals that the effective retrieval modality depends on the reasoning path, while image-set questions remain particularly challenging due to the combined difficulty of retrieving and jointly interpreting multiple images.

Our main contributions are summarized as follows:
\begin{itemize}
    \item We introduce \textbf{CrossModalQA}, an open-domain multimodal RAG benchmark comprising 1,863 question--answer pairs over 4,987 Wikipedia articles and 4,431 Wikimedia Commons images.

    \item We define five complementary cross-modal reasoning paths and construct multi-hop questions through multimodal knowledge graph-guided subgraph sampling, with explicit supporting evidence for fine-grained retrieval and generation evaluation.

    \item Extensive experiments reveal that existing multimodal RAG systems struggle with complete evidence retrieval and multi-image reasoning. Further analysis confirms that retrieval completeness is the primary factor determining end-to-end answer accuracy.
\end{itemize}

\section{Related Work}

\subsection{Multimodal Retrieval-Augmented Generation}
Retrieval-augmented generation grounds parametric language models in external memory, allowing retrieved evidence to support knowledge-intensive prediction and enabling knowledge updates without retraining the full generator. Early systems couple dense passage retrieval with sequence generation, while later architectures improve evidence aggregation and scale retrieval to much larger corpora~\citep{lewis2020retrieval, guu2020retrieval, borgeaud2022improving}. Multimodal RAG extends this paradigm to corpora containing text passages, standalone images, and aligned image--text items. A typical system first constructs retrievable units from each modality. Unified designs map text and images into a shared vision--language space~\citep{radford2021learning}, while modular designs maintain modality-specific indexes. The retriever selects candidate evidence, a reranker refines its relevance, and an MLLM generates the answer from the query together with the selected context~\citep{chen2022murag, hu2023reveal, penamakuri2023retvqa}. Recent systems increasingly separate retrieval from generation and employ hierarchical search, cross-modal reranking, and evidence filtering to improve coverage while controlling noisy context~\citep{chen2024mllm, bai2025ramqa, tian2025core, yu2025visrag, wang2025vidorag, li2026regionrag, hsiao2026megarag, bu2025query, yuan2026mkg, liu2025benchmarking}. Recently, agentic variants further decompose complex questions, select retrieval tools, and adapt later searches to intermediate results~\citep{liu2025hm}. These developments have made multimodal RAG an increasingly important technology for generative AI.

\subsection{Multimodal RAG Benchmarks}
Whereas early benchmarks often reduced the task to ranking a small set of candidate images within a narrowly curated corpus~\citep{talmor2021multimodalqa, chang2022webqa}, recent benchmarks are built around substantially larger, more heterogeneous, and semantically richer collections~\citep{mensink2023encyclopedic, liu2025benchmarking}. They usually contain a large corpus composed of images, text articles, and visual documents. This shift raises the bar from local candidate selection to genuine multimodal retrieval at scale.
Such benchmarks
evaluate whether models can retrieve relevant visual or textual evidence, suppress distractors, integrate multiple evidence items, and exploit external knowledge beyond their parametric memory~\citep{chen2023can, hu2025mrag, li2025benchmarking, wu2025visual, yu2025mramg}. However, they do not clearly distinguish different reasoning paths with regard to modalities. Moreover, most existing benchmarks focus on a shallow one- or two-hop evidence chains. Although some benchmarks may contain multi-hop questions, reasoning-path diversity and hop complexity are rarely controlled jointly and systematically. In contrast, CrossModalQA explicitly covers all five modality-reasoning categories and introduces multi-hop questions that require models to retrieve and compose evidence across multiple intermediate steps, enabling fine-grained evaluation across the dimensions of retrieval breadth and reasoning depth.

\begin{table*}[t]
    \centering
    \renewcommand{\arraystretch}{1.12}
    \setlength{\tabcolsep}{3.5pt}

    \begin{tabular}{@{}lccccc@{}}
        \toprule
        \textbf{Method}
        & \multicolumn{5}{c}{\textbf{Question Category}} \\
        \cmidrule(lr){2-6}
        & \makecell[c]{\textbf{V-to-T}}
        & \makecell[c]{\textbf{T-to-V}}
        & \makecell[c]{\textbf{V-to-T-to-V}}
        & \makecell[c]{\textbf{Multi-image}\\
                       \textbf{Intersection}}
        & \makecell[c]{\textbf{Image-set}} \\
        \midrule

        \multicolumn{6}{l}{\textit{Text-only Questions}} \\
        MultiModalQA\footnotesize~\citep{talmor2021multimodalqa}
        & \cmark & \cmark & \xmark & \cmark & \cmark \\

        WebQA\footnotesize~\citep{chang2022webqa}
        & \cmark & \cmark & \xmark & \xmark & \xmark \\

        RetVQA\footnotesize~\citep{penamakuri2023retvqa}
        & \xmark & \cmark & \xmark & \cmark & \xmark \\

        Visual-RAG\footnotesize~\citep{wu2025visual}
        & \xmark & \cmark & \xmark & \xmark & \xmark \\

        MRAMG-Bench\footnotesize~\citep{yu2025mramg}
        & \xmark & \cmark & \xmark & \xmark & \xmark \\

        \midrule
        \multicolumn{6}{l}{\textit{Text-and-image Questions}} \\
        InfoSeek\footnotesize~\citep{chen2023can}
        & \cmark & \xmark & \xmark & \xmark & \xmark \\

        Encyclopedic VQA\footnotesize~\citep{mensink2023encyclopedic}
        & \cmark & \xmark & \xmark & \xmark & \xmark \\

        MRAG-Bench\footnotesize~\citep{hu2025mrag}
        & \cmark & \xmark & \cmark & \xmark & \xmark \\

        Dyn-VQA\footnotesize~\citep{li2025benchmarking}
        & \cmark & \xmark & \cmark & \xmark & \xmark \\

        \midrule
        \multicolumn{6}{l}{\textit{Mixed Questions}} \\
        M$^2$RAG\footnotesize~\citep{liu2025benchmarking}
        & \cmark & \cmark & \xmark & \xmark & \xmark \\

        \midrule
        \multicolumn{6}{l}{\textit{Text-only Questions}} \\
        \textbf{CrossModalQA}
        & \cmark
        & \cmark
        & \cmark
        & \cmark
        & \cmark \\

        \bottomrule
    \end{tabular}

    \caption{The five categories in CrossModalQA. ``V'' denotes vision while ``T'' denotes text. For text-only questions, V-to-T and V-to-T-to-V denote that the question describes visual features. For text-and-image questions, V-to-T and V-to-T-to-V denote that the question has attached images.}
    \label{tab:five_modes}
\end{table*}

\begin{figure*}[ht]
    \centering
    \includegraphics[width=0.95\textwidth]{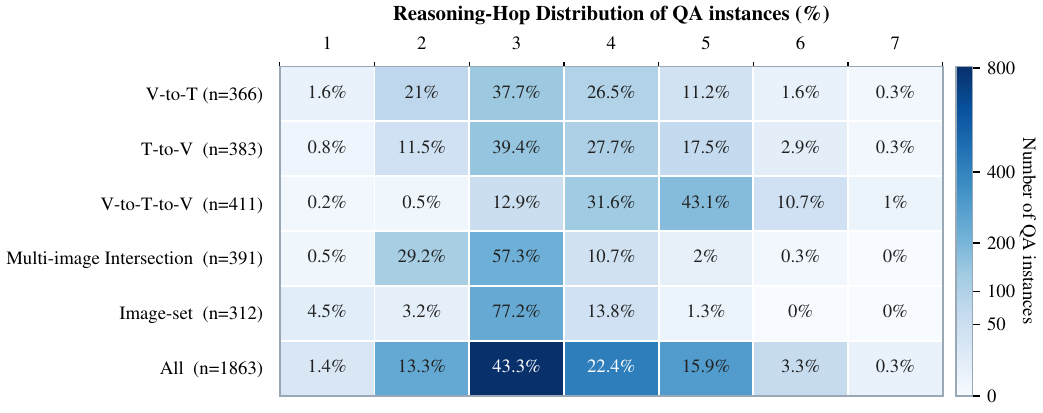}
    \caption{Distribution of QA instances by reasoning-hop count across the five question modes. Each cell reports the row-wise percentage of QA instances, while color intensity represents the corresponding absolute count. Most questions require three or four reasoning hops. Specifically, the Multi-image Intersection and Image-set mode exhibits a greater concentration of two- and three-hop questions, consistent with their design goal of evaluating a model’s ability to retrieve multiple relevant images from a large image corpus, with less emphasis on long reasoning chains. The average number of reasoning hops is 3.50.}
    \label{fig:hop}
\end{figure*}

\section{CrossModalQA}
\subsection{Dataset Overview}
We introduce \textbf{CrossModalQA}, a multimodal RAG benchmark containing 1,863 question-answer pairs constructed from 4,987 Wikipedia articles and 4,431 images collected from Wikimedia Commons. 
The questions in CrossModalQA are categorized according to their reasoning paths over visual and textual evidence. The questions also require multi-hop reasoning over data from different modalities. 
To the best of our knowledge, CrossModalQA is the first multimodal RAG benchmark that covers all five cross-modality reasoning-path categories.

\subsection{Reasoning-path Taxonomy}
\label{subsec:taxonomy}
Answering each question requires the solver to ground various entities in the visual or textual modality and reason over the facts connecting them before arriving at a final answer, which may itself be a textual fact or a visual attribute. Based on the structure of this reasoning process, we organize CrossModalQA into five modes:
\paragraph{Vision-to-text.}
In this mode, the question first describes visual evidence that can be used to identify an entity. After grounding this visually specified entity, the solver must follow textual facts to derive an entity. The final answer is the name or a non-visual attribute of this entity.

\paragraph{Text-to-vision.}
The solver first uses textual facts in the KG to infer the target entity, and then inspects the corresponding image to answer a concrete visual attribute, such as color, shape, material appearance, layout, or visible wording.

\paragraph{Vision-to-text-to-vision.}
Questions in this mode require two visual grounding steps connected by textual reasoning. The solver identifies a source entity from visual clues, follows one or more KG facts to a related target entity, and finally answers by inspecting the target entity's image.

\paragraph{Multi-image intersection reasoning.}
This mode involves multiple visually grounded entities that form different branches of a connected subgraph. The solver must recognize the relevant entities from their images, trace their textual relations, and find a shared answer, such as a common class, location, or entity satisfying the intersection of these constraints.

\paragraph{Image-set reasoning.}
In this mode, the KG defines a candidate set of related entities, and the solver must compare several associated images to determine which candidates satisfy a visual condition. The answer may be a single entity or multiple entities, making this mode closer to set-level multimodal reasoning than pairwise entity grounding.

\subsection{Multi-hop Multimodal Reasoning}
Existing multimodal retrieval-augmented QA benchmarks are often dominated by short reasoning chains, typically requiring only one or two retrieval or reasoning steps. In contrast, CrossModalQA is designed to emphasize multi-hop multimodal reasoning across modalities. Each question is grounded in a retained multimodal subgraph, and solving it requires composing evidence across image nodes, entity nodes, and textual relations.

The reasoning paths are instantiated on a multimodal knowledge graph, where both entities and images are represented as nodes. Edges include textual relations between entities as well as image-entity links. Under this formulation, moving from an image to its corresponding entity, from one entity to another entity, or from an entity to an associated image is treated uniformly as one reasoning hop.

Formally, for each question, let $\mathcal{G}_q=(\mathcal{V}_q,\mathcal{E}_q)$ denote the retained multimodal subgraph used to support the reasoning process. Let $a \in \mathcal{V}_q$ be the node corresponding to the selected answer, where $a$ can be either an entity node or an image node. When $a$ is an image node, the answer is normally about the visual feature of the entity node connected to $a$. We define the reasoning hop of the question as
\[
h(q)=\max_{v\in \mathcal{V}_q} d_{\mathcal{G}_q}(v,a),
\]
where $d_{\mathcal{G}_q}(v,a)$ is the shortest-path distance from node $v$ to the answer node $a$ within $\mathcal{G}_q$. This definition measures the largest number of reasoning steps required to connect any retained evidence node to the final answer.

Table~\ref{fig:hop} shows the number of reasoning hops in CrossModalQA. The average number of reasoning hops is 3.50, and over 40\% of the questions require four or more reasoning hops.
Most questions require three or
four reasoning hops, which is huge improvement compared with existing benchmarks. Specifically, the Multi-image Intersection and Image-set mode exhibits a greater concentration of two- and
three-hop questions, consistent with their design goal of evaluating a model’s ability to retrieve multiple relevant images from a
large image corpus, with less emphasis on long reasoning chains.

\subsection{Multimodal KG-Guided QA Generation}

We construct CrossModalQA through a four-stage pipeline: (1) domain-targeted article selection, (2) knowledge graph (KG) extraction, (3) multimodal grounding via Wikidata and Wikimedia Commons, and (4) subgraph-conditioned QA generation with quality control.

\paragraph{Article Selection.} We source Wikipedia articles from FineWiki~\citep{penedo2025finewiki}, a large-scale corpus of parsed Wikipedia dumps snapshotted in August 2025. To ensure that visual evidence is essential, rather than incidental, to answering each question, we restrict candidate articles to domains with strong visual groundedness: animals and animal taxa, plants and plant taxa, buildings and structures, paintings, sculptures, and other cultural artifacts, as well as entities characterized by a distinctive flag or logo.

\paragraph{Knowledge Graph Extraction.} From the selected articles, we extract 
\texttt{(subject, relation, object)} triples to instantiate a textual KG spanning these domains.

\paragraph{Multimodal Grounding.} We link each KG entity to its corresponding Wikidata item and retrieve representative images via Wikidata's image-related properties (e.g., P18 \textit{image}), hosted on Wikimedia Commons. As these image associations are often noisy, we further employ a vision-language model
to verify entity-image consistency, discarding images that fail to faithfully depict their associated entity. Validated entity-image pairs are then linked to their corresponding KG nodes, yielding a multimodal knowledge graph (MMKG) in which each node is optionally augmented with a canonical image alongside its textual attributes.

\paragraph{QA Generation and Quality Control.} For each instance, we sample a connected subgraph from the MMKG to serve as grounding context. Conditioned on this subgraph, an LLM generates a question-answer pair instantiating one of a set of predefined question modes. All generated QA pairs are then subjected to a two-stage filtering process combining rule-based consistency checks with LLM-based verification. Only instances that conform to their intended question mode and necessitate multimodal evidence for resolution are retained in the final benchmark.

\subsection{Evaluation Metrics}

\subsubsection{Retrieval performance}
For each question $q \in \mathcal{Q}$, let $G_q$ denote the set of gold evidence items and let $R_q^{(k)}$ denote the top-$k$ retrieved items. Depending on the retriever, evidence items can be images, text articles, or their union. We evaluate retrieval with Recall@k, Precision@k, and Hit@k for $k \in \{1,5,10\}$:
\[
\mathrm{Recall@}k(q) =
\frac{|R_q^{(k)} \cap G_q|}{|G_q|}, 
\mathrm{Precision@}k(q) =
\frac{|R_q^{(k)} \cap G_q|}{k}, 
\mathrm{Hit@}k(q) =
\mathbb{I}\left(|R_q^{(k)} \cap G_q| > 0\right),
\]
where $\mathbb{I}[\cdot]$ denotes the indicator function. We report macro-averaged scores over all questions:
\[
\mathrm{Metric@}k =
\frac{1}{|\mathcal{Q}|}
\sum_{q \in \mathcal{Q}} \mathrm{Metric@}k(q).
\]
Intuitively, Recall@k measures the fraction of gold evidence recovered, Precision@k measures the fraction of retrieved items that are relevant, and Hit@k measures whether at least one gold evidence item appears in the top-$k$ results.

\subsubsection{Generation performance} 
\label{subsec:eval}
We evaluate generation quality at the level of atomic answer units. For each question, the reference answer and the model response are represented as sets of atomic facts, from which we compute true positives (TP), false positives (FP), and false negatives (FN). 
The \emph{factual correctness} is measured by the atom-level F1 score:

\[
F_1
=
\frac{2\operatorname{TP}}
{2\operatorname{TP}+\operatorname{FP}+\operatorname{FN}}.
\]
We further report atom-level exact match (EM), 
defined as

\[
\mathrm{EM}
=
\mathbb{I}\!\left(\mathrm{FP}=0 \,\land\, \mathrm{FN}=0\right),
\]
i.e., a response is credited only if it contains no unsupported atoms and omits no reference atoms.

\section{Experiments}

\begin{table}[ht]
\centering
\small
\renewcommand{\arraystretch}{1.1}
\resizebox{\linewidth}{!}{%
\begin{tabular}{@{}l*{6}{cc}@{}}
\toprule
\textbf{Method}
& \multicolumn{2}{c}{\textbf{V to T}}
& \multicolumn{2}{c}{\textbf{T to V}}
& \multicolumn{2}{c}{\textbf{V to T to V}}
& \multicolumn{2}{c}{\makecell[c]{\textbf{Multi-image}\\\textbf{Intersection}}}
& \multicolumn{2}{c}{\textbf{Image-set}}
& \multicolumn{2}{c}{\textbf{Overall}} \\
\cmidrule(lr){2-3}
\cmidrule(lr){4-5}
\cmidrule(lr){6-7}
\cmidrule(lr){8-9}
\cmidrule(lr){10-11}
\cmidrule(lr){12-13}

& EM & F1
& EM & F1
& EM & F1
& EM & F1
& EM & F1
& EM & F1 \\
\midrule

GPT 4o
& 20.2 & 20.9
& 29.8 & 32.1
& 37.2 & 39.3
& 21.7 & 21.5
& 8.3 & 23.9
& 24.3 & 27.6 \\

GPT 4o + Gold texts
& 75.7 & 77.5
& 31.1 & 33.4
& 39.4 & 41.1
& 81.1 & 80.9
& 17.0 & 44.1
& 49.8 & 54.2 \\

GPT 4o + Gold images
& 30.9 & 31.8
& \textbf{66.6} & \textbf{68.0}
& \textbf{74.7} & \textbf{75.8}
& 43.5 & 43.6
& \textbf{29.2} & 52.2
& 50.2 & 54.7 \\

GPT 4o + Gold texts \& images
& \textbf{77.3} & \textbf{79.1}
& 64.2 & 66.4
& 70.1 & 70.8
& \textbf{82.9} & \textbf{83.4}
& \textbf{29.2} & \textbf{56.6}
& \textbf{66.1} & \textbf{70.3} \\

\bottomrule
\end{tabular}%
}
\caption{Performance of GPT-4o under different oracle-evidence settings. Gold texts and Gold images denote the provision of ground-truth evidence from the corresponding modality. We report Exact Match (EM) and micro-F1 for each of the five question modes, together with their overall
average.}
\label{tab:ablation_results}
\end{table}

\subsection{Experimental Setup}
\label{sec:experimental_setup}

\paragraph{Baselines settings.}
We evaluate two representative multimodal RAG baselines. M$^2$RAG~\citep{liu2025benchmarking} retrieves textual articles and images from the benchmark corpus, while mKG-RAG~\citep{yuan2026mkg} performs structured retrieval over a multimodal knowledge graph. The original query-aware retriever of mKG-RAG requires an image--question pair. Since CrossModalQA contains text-only questions, we evaluate its two applicable text-query variants: text-to-image (T2I) and text-to-text (T2T) retrieval. We further include GPT-4o mini and GPT-4o~\citep{hurst2024gpt} under a closed-book setting, where each model answers solely from its parametric knowledge.
To ensure a controlled comparison, we use GPT-4o as the answer generator for all RAG methods. More details are provided in the appendix.

\paragraph{Ablation study settings.}
For oracle analysis, we evaluate GPT-4o under four settings: with no external knowledge, gold text articles, gold images, and the combination of gold text articles and images. 

\paragraph{Evaluation.}
Retrieval performance is measured using Recall@5, Recall@10, Precision@5, and Hit@5. Answer quality is measured using exact match (EM) and micro-F1. We report results for each question mode and the complete benchmark.

\begin{table*}[ht]
\centering
\small
\renewcommand{\arraystretch}{1.1}
\resizebox{\linewidth}{!}{
\begin{tabular}{@{}l*{6}{cc}@{}}
\toprule
\textbf{Method}
& \multicolumn{2}{c}{\textbf{V to T}}
& \multicolumn{2}{c}{\textbf{T to V}}
& \multicolumn{2}{c}{\textbf{V to T to V}}
& \multicolumn{2}{c}{\makecell[c]{\textbf{Multi-image}\\\textbf{Intersection}}}
& \multicolumn{2}{c}{\textbf{Image-set}}
& \multicolumn{2}{c}{\textbf{Overall}} \\
\cmidrule(lr){2-3}
\cmidrule(lr){4-5}
\cmidrule(lr){6-7}
\cmidrule(lr){8-9}
\cmidrule(lr){10-11}
\cmidrule(lr){12-13}

& EM & F1
& EM & F1
& EM & F1
& EM & F1
& EM & F1
& EM & F1 \\
\midrule




Vanilla GPT 4o mini
& 13.7 & 14.6
& 27.4 & 28.8
& 30.4 & 31.7
& 17.4 & 17.4
& 6.7 & 22.7
& 19.8 & 23.2    \\

Vanilla GPT 4o
& 20.2 & 20.9
& 29.8 & 32.1
& 37.2 & 39.3
& 21.7 & 21.5
& 8.3 & 23.9
& 24.3 & 27.6 \\
\midrule


M$^2$RAG \footnotesize\citep{liu2025benchmarking}
& 11.7 & 28.2
& 17.8 & 27.7
& 15.1 & 22.5
& 19.7 & 25.8
& 4.8 & 33.2
& 14.2 & 27.5 \\


mKG-RAG (T2I)\footnotesize~\citep{yuan2026mkg}
& 14.8 & 15.3
& 21.7 & 21.5
& 28.0 & 28.4
& 20.7 & 20.3
& 6.7 & 15.6
& 19.0 & 20.1 \\

mKG-RAG (T2T)\footnotesize~\citep{yuan2026mkg}
& 21.6 & 20.8
& 14.9 & 15.5
& 27.7 & 28.0
& 21.7 & 20.6
& 7.4 & 24.3
& 19.2 & 22.1 \\

\bottomrule
\end{tabular}
}
\caption{Exact-match and F1 scores (\%).}
\label{tab:baseline_results}
\end{table*}

\begin{table*}[t]
\centering
\small
\setlength{\tabcolsep}{1.2pt}
\renewcommand{\arraystretch}{1.15}

\resizebox{\linewidth}{!}{
\begin{tabular}{@{}l*{12}{c}@{}}
\toprule
\textbf{Method}
& \multicolumn{4}{c}{\textbf{V to T}}
& \multicolumn{4}{c}{\textbf{T to V}}
& \multicolumn{4}{c}{\textbf{V to T to V}} \\
\cmidrule(lr){2-5}
\cmidrule(lr){6-9}
\cmidrule(lr){10-13}
& \textbf{R@5} & \textbf{R@10} & \textbf{P@5} & \textbf{H@5}
& \textbf{R@5} & \textbf{R@10} & \textbf{P@5} & \textbf{H@5}
& \textbf{R@5} & \textbf{R@10} & \textbf{P@5} & \textbf{H@5} \\
\midrule
\multicolumn{13}{l}{\textit{ Image retrieval}}\\

mKG-RAG (T2I)\footnotesize~\citep{yuan2026mkg}
& 51.0 & 60.8 & 12.0 & 56.8
& 45.3 & 52.1 & 9.1 & 45.4
& 32.8 & 42.3 & 13.4 & 55.5  \\
\multicolumn{13}{l}{\textit{ Multimodal retrieval}}\\
M$^2$RAG \footnotesize\citep{liu2025benchmarking}
& 36.9 & 45.1 & 19.3 & 70.2
& 49.0 & 55.3 & 24.4 & 85.1
& 23.9 & 30.0 & 16.7 & 58.6 \\
\midrule
\textbf{Method}
& \multicolumn{4}{c}{\makecell[c]{\textbf{Multi-image}\\\textbf{Intersection}}}
& \multicolumn{4}{c}{\textbf{Image-set}}
& \multicolumn{4}{c}{\textbf{Overall}} \\
\cmidrule(lr){2-5}
\cmidrule(lr){6-9}
\cmidrule(lr){10-13}
& \textbf{R@5} & \textbf{R@10} & \textbf{P@5} & \textbf{H@5}
& \textbf{R@5} & \textbf{R@10} & \textbf{P@5} & \textbf{H@5}
& \textbf{R@5} & \textbf{R@10} & \textbf{P@5} & \textbf{H@5} \\
\midrule
\multicolumn{13}{l}{\textit{ Image retrieval}}\\


mKG-RAG (T2I)\footnotesize~\citep{yuan2026mkg}
& 37.4 & 45.8 & 17.2 & 64.7
& 21.8 & 29.9 & 16.5 & 53.5
& 38.1 & 46.6 & 13.6 & 55.3 \\
\multicolumn{13}{l}{\textit{ Multimodal retrieval}}\\
M$^2$RAG \footnotesize\citep{liu2025benchmarking}
& 23.4 & 29.5 & 17.7 & 59.1
& 27.8 & 32.0 & 28.3 & 85.9
& 32.2 & 38.4 & 20.9 & 71.0 \\
\bottomrule
\end{tabular}}
\caption{Retrieval performance. R@5, R@10, P@5, and H@5 denote Recall@5, Recall@10, Precision@5, and Hit@5, respectively. mKG-RAG (T2I) is image-only retrieval, while M$^2$RAG retrieves both images and text rticles.}
\label{tab:retrieval_performance}
\end{table*}

\subsection{Ablation Study}

We here examine the \textbf{necessity of multimodal evidence} in answering CrossModalQA questions.
Table~\ref{tab:ablation_results} evaluates GPT-4o under different oracle-evidence settings. The no-retrieval setting and both single-modality settings remain substantially below the joint text--image oracle in overall answer accuracy. This consistent gap shows that the questions in CrossModalQA require complementary evidence from both images and text articles. Successful reasoning paths therefore depend on retrieving both modalities.


Notably, providing both gold modalities does not consistently outperform supplying only the gold images. This result suggests that, For T-to-V and V-to-T-to-V questions, the target image conveys an entity intended to be recovered through preceding reasoning steps. The result that providing gold images yields higher accuracy than providing gold texts and images together suggests directly supplying the gold image reveals this entity in advance and shortens the required evidence path. We therefore interpret the observed advantage as an oracle shortcut created by early access to the target visual evidence.

\subsection{Result Analysis}

Table~\ref{tab:baseline_results} shows that the evaluated multimodal RAG methods struggle to support retrieval and reasoning across multiple articles and images. We attribute their limited performance to the following factors:

\paragraph{Retrieval errors severely impair answer accuracy.}
Failed retrieval introduces irrelevant evidence while omitting information required by the reasoning chain. The resulting noise misleads the generator and causes every evaluated multimodal RAG baseline to underperform GPT-4o using only its parametric knowledge.

\paragraph{A single relevant retrieval result provides insufficient evidence for complex questions.}
M$^2$RAG achieves an high Hit@5 yet obtains a low EM, while its recall remains limited. Hit@5 measures whether the retrieved set contains any relevant item, whereas multimodal multi-hop questions require multiple complementary evidence items. This gap shows that partial retrieval rarely provides sufficient support for completing the full reasoning chain.

\par\medskip
The results in Table~\ref{tab:ablation_results},~\ref{tab:baseline_results} and~\ref{tab:retrieval_performance} also lead to following observations:

\paragraph{The effective retrieval modality aligns with the task-specific reasoning direction.}
The oracle and baseline results in Table~\ref{tab:ablation_results} and~\ref{tab:baseline_results} jointly reveal a correspondence between retrieval modality and question type. 
V-to-T and Multi-image Intersection questions first identify entities from visual features and then reason over textual facts, making them more sensitive to text retrieval.
In contrast, T-to-V and V-to-T-to-V questions are primarily constrained by visual retrieval. These questions seek answers grounded in target images, making their performance highly responsive to accurate image retrieval.

\paragraph{Image-set reasoning remains the most challenging mode.}
Table~\ref{tab:retrieval_performance} shows that existing methods recover only a limited portion of the required image set. Table~\ref{tab:ablation_results} further shows that end-to-end performance remains substantially below that of other question modes even when gold evidence is provided. This result reveal limited model capacity to jointly interpret multiple images and integrate their visual evidence into a complete answer. The incomplete multi-image retrieval and limited multi-image reasoning capability are two compounding bottlenecks.

\paragraph{Accurate retrieval contributes more than generator scaling to answer accuracy.}
Gold evidence produces substantially larger accuracy gains than replacing the generator with a stronger model. This contrast identifies accurate retrieval as the dominant factor in end-to-end performance and emphasizes the importance of recovering the complete evidence required by each reasoning path.

\section{Conclusion}

We introduce \textbf{CrossModalQA}, an open-domain benchmark for evaluating multimodal RAG under diverse cross-modal reasoning structures. CrossModalQA poses complicated questions over a mixed corpus of Wikipedia articles and Wikimedia Commons images and organizes them into five reasoning-path categories, with each instance requiring multi-hop composition of distributed textual and visual evidence. Its multimodal knowledge graph-guided construction retains explicit supporting subgraphs, enabling fine-grained evaluation of retrieval and generation across both reasoning path and reasoning depth.

Experiments with representative multimodal RAG systems reveal substantial limitations in complete evidence retrieval and multi-image reasoning. By jointly covering diverse modality transitions and multi-hop reasoning, CrossModalQA provides a challenging testbed for diagnosing these bottlenecks and advancing reliable multimodal RAG.

\bibliography{reference}
\bibliographystyle{iclr2027_conference}

\appendix



\section{Implementation Details}

\subsection{Domain-Targeted Article Selection}
\label{app:domain-targeted-selection}

This stage curates a pool of candidate articles that are likely to be associated with rich visual content, prior to question construction.

We draw candidate articles from \textbf{FineWiki}~\citep{penedo2025finewiki}, a large-scale Wikipedia corpus released by Hugging Face in 2025. FineWiki contains parsed Wikipedia HTML dumps from August of 2025 covering 325 languages. We focus on the English subset of FineWiki, which has more than 61 million articles in total.

To identify articles from visually rich domains, we prompt GPT-5.4-nano
to assign each candidate article to exactly one of eight categories, shown in Table~\ref{tab:article-domain}. Categories A--G correspond to domains expected to yield visually descriptive content, each capped at a quota to prevent any single domain from dominating the pool. Category H (\emph{Else}) captures all remaining articles and carries no fixed quota. 

We build a huge multimodal knowledge graph using all selected articles. After the full dataset construction pipeline, we discard articles that are not referenced by any retained question, resulting in a final set of 4,987 articles.

\begin{table}[ht]
\centering
\setlength{\tabcolsep}{3pt}
\begin{tabular}{@{}cp{0.70\linewidth}r@{}}
\toprule
\textbf{ID} & \textbf{Category} & \textbf{Quota} \\
\midrule
A & Animals and animal taxa & 2{,}000 \\
B & Plants and plant taxa & 2{,}000 \\
C & Buildings and structures & 2{,}000 \\
D & Paintings & 2{,}000 \\
E & Sculptures & 500 \\
F & Other cultural relics & 1{,}000 \\
G & Entities with a distinctive flag or logo & 500 \\
H & Else & -- \\
\bottomrule
\end{tabular}
\caption{Category taxonomy used for domain-targeted article selection. The quota specifies the number of articles retained in each category. The total number of articles we select is 10,000.}
\label{tab:article-domain}
\end{table}

\subsection{Images Collection}

We first extract atomic KG triples from selected FineWiki articles using GPT-5.1. Entity mentions in the extracted triples are then linked to Wikidata QIDs using deterministic title matching and Wikidata search, with GPT-5.4-mini with web search used to resolve otherwise unresolved cases. For each linked QID, we collect candidate images from Wikidata image-related properties, following the priority order P18 (image), P180 (depicts), P921 (main subject), and P373 (Commons category).
Because P180, P921, and P373 may retrieve images that are only loosely associated with the target entity, we further verify the entity-image pairs using GPT-5.1 with visual input and filter out mismatched images. This additional verification step is used to improve the visual grounding quality of the final dataset.

\begin{figure}[t]
\centering
\includegraphics[width=0.7\columnwidth]{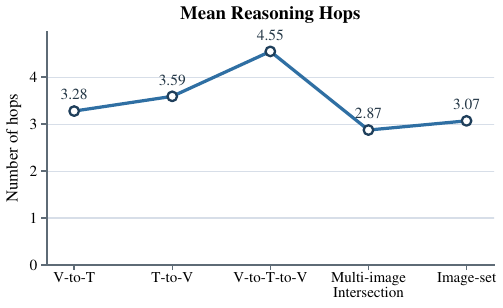} 
\caption{Average reasoning hops of CrossModalQA by mode.}
\label{fig:avg_hop}
\end{figure}

\begin{figure}[t]
\centering
\includegraphics[width=0.65\columnwidth]{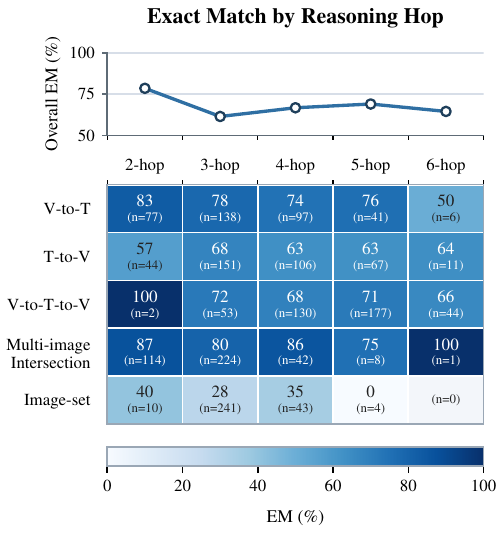} 
\caption{Exact Match (EM) by reasoning hop and QA mode in the \textit{GPT-4o + Gold texts \& images} setting. The line plot shows the overall EM for each reasoning hop, while the heatmap reports the per-mode EM. Values in parentheses indicate the number of QA instances in each cell. We exclude the 1-hop and 7-hop cases because of insufficient sample sizes for reliable evaluation. EM scores are reported on a 0--100 scale and rounded to the nearest integer.}
\label{fig:em_by_hop}
\end{figure}

\begin{figure}[t]
\centering
\includegraphics[width=0.65\columnwidth]{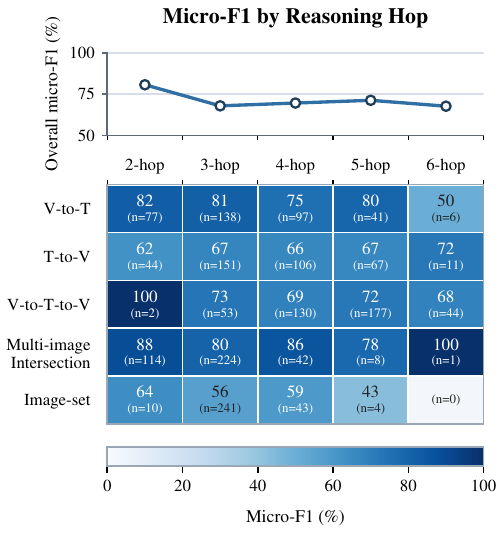} 
\caption{Micro-F1 by reasoning hop and QA mode in the \textit{GPT-4o + Gold texts \& images} setting. The line plot shows the overall Micro-F1 for each reasoning hop, while the heatmap reports the per-mode Micro-F1. Values in parentheses indicate the number of QA instances in each cell. We exclude the 1-hop and 7-hop cases because of insufficient sample sizes for reliable evaluation. F1 scores are reported on a 0--100 scale and rounded to the nearest integer.}
\label{fig:f1_by_hop}
\end{figure}






\subsection{Dataset Construction Details}

\paragraph{Subgraph sampling.}
We construct a multimodal knowledge graph
$\mathcal{G}=(V, E_f \cup E_i)$,
where $E_f$ denotes textual factual triples and
$E_i$ denotes image attachment edges of the form
$(v,\texttt{has\_image},x)$. 

The detailed optimization process is described in Algorithm~\ref{alg:mm-subgraph-sampling}. For each question mode $m$, let
$\theta_m=(L_m,U_m,F_m^{\min},I_m^{\min},I_m^{\max},H_m,Q_m)$ denote its
sampling configuration, where $[L_m,U_m]$ is the total edge budget range,
$F_m^{\min}$ is the minimum number of factual edges,
$[I_m^{\min},I_m^{\max}]$ is the image-entity budget range, $H_m$ is the
randomized factual expansion depth, and $Q_m$ is a lower bound on
the multimodal subgraph hop count. Given target count $T_m$, and attempt multiplier $\alpha$, the sampler draws at most
$\alpha T_m$ candidates. 

\begin{figure*}[t]
\centering
\includegraphics[width=0.99\textwidth]{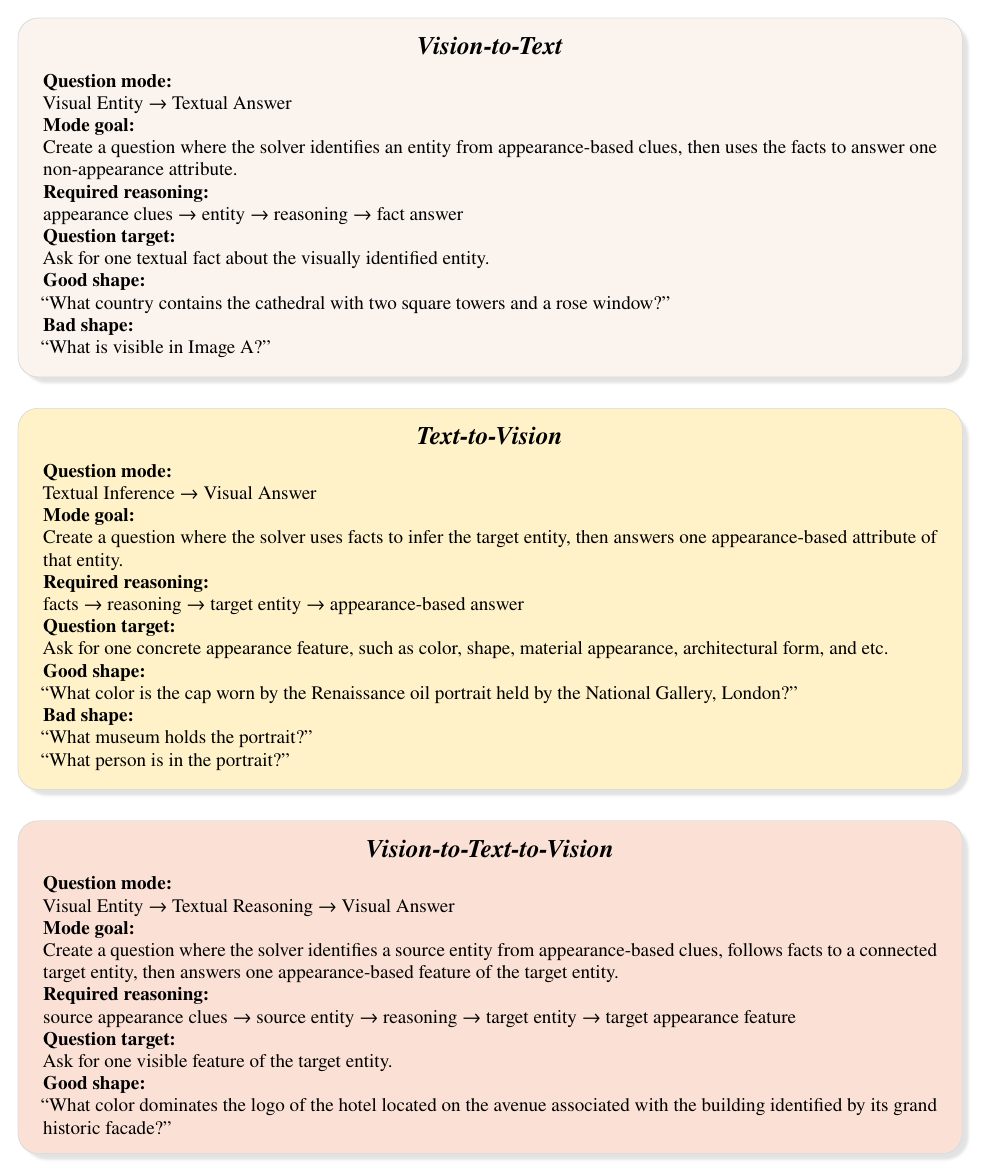} 
\caption{Example of prompts used in question generation stage.}
\label{fig:prompt123}
\end{figure*}

\begin{figure*}[t]
\centering
\includegraphics[width=0.99\textwidth]{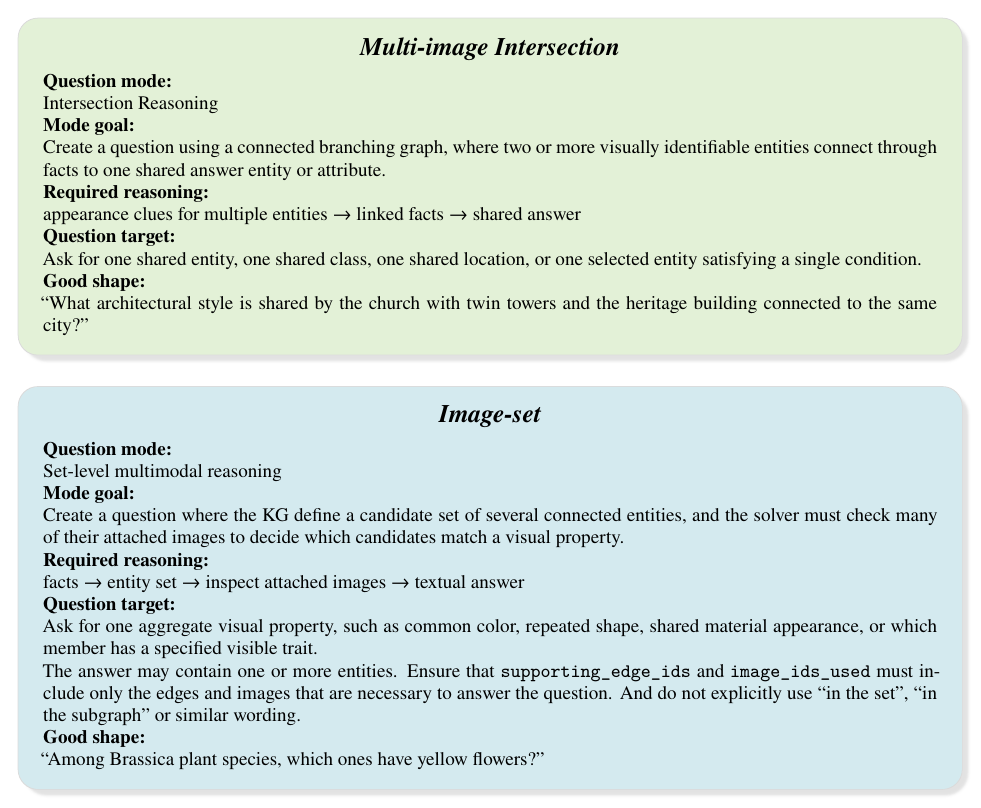} 
\caption{Example of prompts used in question generation stage.}
\label{fig:prompt45}
\end{figure*}

\paragraph{QA Generation}
We employ GPT-5.4 model to generate our QAs according to the subgraph. Figure~\ref{fig:prompt123} and~\ref{fig:prompt45} show the prompts we use.

\begin{algorithm}[t]
\caption{Mode-specific multimodal subgraph sampling}
\label{alg:mm-subgraph-sampling}
\begin{algorithmic}
\REQUIRE multimodal graph $\mathcal{G}=(V, E_f \cup E_i)$, image table
$\mathcal{M}$, mode config $\theta_m$, target count $T_m$, attempt multiplier $\alpha$
\ENSURE sampled subgraph pool $\mathcal{S}_m$

\STATE $\mathcal{S}_m \gets \emptyset$
\STATE $\mathcal{D} \gets \emptyset$ \COMMENT{seen edge-id signatures}
\STATE build undirected factual adjacency $\mathcal{A}_f$ from $E_f$
\STATE $\mathcal{B}(v) \gets \{e \in E_i \mid e.\mathrm{subject}=v
\land e.\mathrm{image\_id}\in\mathcal{M}\}$
\STATE $\mathcal{U} \gets \{v \mid \mathcal{B}(v)\neq\emptyset
\land \mathcal{A}_f(v)\neq\emptyset\}$

\FOR{$a = 1$ \textbf{to} $\alpha T_m$}
    \IF{$|\mathcal{S}_m| = T_m$}
        \STATE \textbf{break}
    \ENDIF
    \STATE sample $B \sim \mathrm{UnifInt}(L_m,U_m)$
    \IF{$B-F_m^{\min}<I_m^{\min}$}
        \STATE \textbf{continue}
    \ENDIF
    \STATE sample $K \sim \mathrm{UnifInt}(I_m^{\min},
    \min(I_m^{\max}, B-F_m^{\min}))$
    \STATE sample seed $s \sim \mathrm{Uniform}(\mathcal{U})$
    \STATE $F \gets \textsc{RandomBFSFacts}(s,\mathcal{A}_f,B-K,H_m)$
    \IF{$|F| < F_m^{\min}$}
        \STATE \textbf{continue}
    \ENDIF
    \STATE $C \gets \{v \in V(F) \mid \mathcal{B}(v)\neq\emptyset\}$
    \IF{$|C| < I_m^{\min}$}
        \STATE \textbf{continue}
    \ENDIF
    \STATE shuffle $C$ and keep the first $\min(K,|C|)$ entities
    \STATE $I \gets$ one random \texttt{has\_image} edge from $\mathcal{B}(v)$
    for each retained entity $v$
    \STATE $S \gets F \cup I$
    \IF{$|S| < L_m$ or $|S| > B$}
        \STATE \textbf{continue}
    \ENDIF
    \STATE $\sigma \gets$ sorted edge-id signature of $S$
    \IF{$\sigma \in \mathcal{D}$}
        \STATE \textbf{continue}
    \ENDIF
    \STATE $\mathcal{D} \gets \mathcal{D} \cup \{\sigma\}$
    \IF{$S$ is disconnected}
        \STATE \textbf{continue}
    \ENDIF
    \IF{$Q_m$ is enabled and the maximum shortest-path distance of $S$,
    including image edges, is smaller than $Q_m$}
        \STATE \textbf{continue}
    \ENDIF
    \STATE $\mathcal{S}_m \gets \mathcal{S}_m \cup \{S\}$
\ENDFOR
\STATE \RETURN $\mathcal{S}_m$
\end{algorithmic}
\end{algorithm}

\subsection{Additional Experimental Settings}

To ensure a fair and controlled comparison across different multimodal RAG methods, we use GPT-4o~\citep{gpt4o} as the answer-generation model for our experiment. By keeping the generator fixed, we minimize performance variations attributable to differences in the underlying language model, thereby allowing the observed results to more directly reflect the effectiveness of each method’s retrieval, context construction, and multimodal evidence integration strategies.

We evaluate an adapted version of mKG-RAG~\citep{yuan2026mkg} on our benchmark. mKG-RAG is a multimodal knowledge-graph-augmented RAG framework that constructs document-level multimodal KGs from text-image knowledge sources and performs two-stage retrieval, first recalling candidate documents and then retrieving query-relevant graph evidence for answer generation. Since CrossModalQA questions are text-only, we cannot provide a query image to the model under test. Therefore, we evaluate two text-conditioned variants: T2T, where the question retrieves textual evidence, and T2I, where the question retrieves image-side evidence from the corpus. 

\begin{table*}[t]
\centering
\small
\renewcommand{\arraystretch}{1.1}
\resizebox{\linewidth}{!}{
\begin{tabular}{@{}l*{6}{cc}@{}}
\toprule
\textbf{Method}
& \multicolumn{2}{c}{\textbf{V to T}}
& \multicolumn{2}{c}{\textbf{T to V}}
& \multicolumn{2}{c}{\textbf{V to T to V}}
& \multicolumn{2}{c}{\makecell[c]{\textbf{Multi-image}\\\textbf{Intersection}}}
& \multicolumn{2}{c}{\textbf{Image-set}}
& \multicolumn{2}{c}{\textbf{Overall}} \\

\cmidrule(lr){2-3}
\cmidrule(lr){4-5}
\cmidrule(lr){6-7}
\cmidrule(lr){8-9}
\cmidrule(lr){10-11}
\cmidrule(lr){12-13}

& EM & F1
& EM & F1
& EM & F1
& EM & F1
& EM & F1
& EM & F1 \\
\midrule

GPT-4o mini
& 13.7 & 14.6
& 27.4 & 28.8
& 30.4 & 31.7
& 17.4 & 17.4
& 6.7 & 22.7
& 19.8 & 23.2    \\

GPT-4o mini + Gold texts
& 71.3 & \textbf{74.0}
& 23.5 & 26.8
& 31.4 & 34.0
& 73.7 & \textbf{75.1}
& 14.7 & 42.3
& 43.7 & 49.1 \\

GPT-4o mini + Gold images
& 23.5 & 26.1
& \textbf{65.0} & \textbf{65.9}
& \textbf{66.4} & \textbf{66.7}
& 33.0 & 33.2
& 21.8 & 49.7
& 43.2 & 48.8 \\

GPT-4o mini + Gold texts \& images
& \textbf{71.6} & 73.4
& 59.8 & 60.9
& 59.4 & 61.1
& \textbf{74.2} & \textbf{75.1}
& \textbf{22.4} & \textbf{52.3}
& \textbf{58.8} & \textbf{63.5} \\

\bottomrule
\end{tabular}
}
\caption{Performance of GPT-4o mini under different oracle-evidence settings. Gold texts and Gold images denote the provision of ground-truth evidence from the corresponding modality. We report Exact Match (EM) and micro-F1 for each of the five question modes, together with their overall
average.}
\label{tab:gpt4omini_results}
\end{table*}



\section{Evaluation Metrics}

\subsection{Retrieval performance.}
For each question $q \in \mathcal{Q}$, the benchmark provides a set of gold
evidence items $G_q$. An evidence item is identified by its corpus-level item
identifier. Depending on the retrieval setting, the evidence universe may
contain only text articles, only images, or the union of text and image items.
All metrics are computed in the corresponding evidence universe. For example,
an image retriever is evaluated against gold image evidence, whereas a
multimodal retriever is evaluated against the union of gold text and image
evidence.

Let
\[
R_q = (r_{q,1}, r_{q,2}, \ldots)
\]
denote the ranked list returned by a retriever for question $q$. Let $R_q^{(k)}$ be the set of the first $k$ unique retrieved items.

For each question, we compute:
\[
\mathrm{Recall@}k(q)
=
\frac{|R_q^{(k)} \cap G_q|}{|G_q|},
\]
\[
\mathrm{Precision@}k(q)
=
\frac{|R_q^{(k)} \cap G_q|}{k},
\]
\[
\mathrm{Hit@}k(q)
=
\mathbb{I}\!\left(|R_q^{(k)} \cap G_q| > 0\right),
\]
for $k \in \{5,10\}$. Recall@k measures how much of the gold evidence is
covered by the top-$k$ retrieval results. Precision@k measures how much of the
retrieved top-$k$ list is relevant. Hit@k ignores the number of recovered gold
items and only checks whether at least one gold evidence item is retrieved.

We report macro-averaged retrieval scores over questions:
\[
\mathrm{Recall@}k
=
\frac{1}{|\mathcal{Q}|}
\sum_{q \in \mathcal{Q}}
\frac{|R_q^{(k)} \cap G_q|}{|G_q|},
\]
\[
\mathrm{Precision@}k
=
\frac{1}{|\mathcal{Q}|}
\sum_{q \in \mathcal{Q}}
\frac{|R_q^{(k)} \cap G_q|}{k},
\]
\[
\mathrm{Hit@}k
=
\frac{1}{|\mathcal{Q}|}
\sum_{q \in \mathcal{Q}}
\mathbb{I}\!\left(|R_q^{(k)} \cap G_q| > 0\right).
\]
Thus, each question contributes equally to the final retrieval score, regardless
of how many gold evidence items it has. This avoids giving disproportionate
weight to questions with larger evidence sets.



\subsection{Generation performance.}

We evaluate generation quality at the level of atomic answer units. For each
question $q$, the reference answer is represented as a set of gold atomic facts.
The model response is atomized into candidate answer atoms, and a judge assigns
each atom-level decision into true positives, false positives, and false
negatives. Let $\mathrm{TP}_q$, $\mathrm{FP}_q$, and $\mathrm{FN}_q$ denote the
numbers of true-positive, false-positive, and false-negative atoms for question
$q$, respectively.

For a single question, atom-level precision, recall, and F1 are computed as
\[
P_q
=
\frac{\mathrm{TP}_q}{\mathrm{TP}_q+\mathrm{FP}_q},
\qquad
R_q
=
\frac{\mathrm{TP}_q}{\mathrm{TP}_q+\mathrm{FN}_q},
\]
\[
F_{1,q}
=
\frac{2P_qR_q}{P_q+R_q}
=
\frac{2\mathrm{TP}_q}
{2\mathrm{TP}_q+\mathrm{FP}_q+\mathrm{FN}_q}.
\]

We also compute atom-level exact match for each question:
\[
\mathrm{EM}_q
=
\mathbb{I}\!\left(\mathrm{FP}_q = 0 \,\land\, \mathrm{FN}_q = 0\right).
\]
Thus, a response receives exact-match credit only if it contains no unsupported
candidate atoms and omits no gold atoms.

Over the full evaluated set $\mathcal{Q}$, we report dataset-level EM as the average of per-question exact
matches:
\[
\mathrm{EM}
=
\frac{1}{|\mathcal{Q}|}
\sum_{q \in \mathcal{Q}}
\mathbb{I}\!\left(\mathrm{FP}_q = 0 \,\land\, \mathrm{FN}_q = 0\right).
\]

For factual correctness, we report micro-averaged atom-level F1. Let
\[
\mathrm{TP}_{\Sigma}
=
\sum_{q \in \mathcal{Q}}
\mathrm{TP}_q,
\quad
\mathrm{FP}_{\Sigma}
=
\sum_{q \in \mathcal{Q}}
\mathrm{FP}_q,
\quad
\mathrm{FN}_{\Sigma}
=
\sum_{q \in \mathcal{Q}}
\mathrm{FN}_q .
\]
The micro F1 score is then
\[
F_{1,\mathrm{micro}}
=
\frac{2\mathrm{TP}_{\Sigma}}
{
2\mathrm{TP}_{\Sigma}
+\mathrm{FP}_{\Sigma}
+\mathrm{FN}_{\Sigma}
}.
\]

\section{Additional Results}
\paragraph{Average number of hops by mode.} CrossModalQA is a benchmark consisting of multi-hop and multimodal questions. Figure~\ref{fig:avg_hop} shows the average number of reasoning hops by question modes. Across the entire benchmark, the average number of reasoning hops required is 3.50.

\paragraph{Answer correctness by the number of hops.}
Figure~\ref{fig:em_by_hop} and~\ref{fig:f1_by_hop} provide a fine-grained breakdown of model performance by reasoning hop and QA mode. 
We observe that 3-hop questions yield lower performance than both 2-hop and 4-hop questions, likely because most difficult question mode, Image-set questions, require 3 hops. This finding suggests that hop count alone does not fully account for question difficulty; the QA modality also plays an important role.

\paragraph{Performance by GPT-4o mini.} We provide additional oracle-evidence setting results produced by GPT-4o mini, a model smaller than GPT-4o.
Table~\ref{tab:gpt4omini_results} evaluates GPT-4o mini under different oracle-evidence settings. Vision-to-Text and Multi-image Intersection questions are more sensitive to text retrieval, while Text-to-Vision and Vision-to-Text-to-Vision questions are more sensitive to image retrieval. The results are consistent with those obtained using GPT-4o.

\section{Frequently Asked Questions}

\subsection{Why Text-only Questions}

A substantial line of multimodal retrieval research assumes an image-conditioned query, in which a reference image serves as the primary retrieval anchor~\cite{wu2024visual, yuan2026mkg}. This assumption, however, does not always hold in practical applications: users may recognize or describe the visual characteristics of an object without having access to an image of it. Moreover, retrieving the correct visual evidence from a fine-grained linguistic description remains challenging for existing vision--language models \cite{wu2025visual, wu2026visret}. We therefore formulate all questions using text alone. This setting enables us to evaluate whether multimodal RAG systems can translate textual descriptions of visual attributes into effective cross-modal retrieval and subsequently reason over the retrieved multimodal evidence. It also prevents systems from relying on a provided image as a direct retrieval shortcut, thereby offering a more stringent and practically relevant assessment of multimodal retrieval and reasoning capabilities.

\subsection{Why Use a Multimodal KG for QA Generation}
We use a multimodal KG as an intermediate representation because it makes QA generation evidence-grounded, controllable, and auditable. The graph provides explicit textual fact edges and image edges, allowing us to sample connected multimodal evidence structures with controlled edge budgets, image counts, and hop constraints. This design encourages multi-hop multimodal reasoning while preserving traceability: each generated question can be linked back to supporting edges and images, enabling automatic statistics, verification, and filtering. Direct generation from raw documents or images would provide less control over reasoning complexity and weaker guarantees about evidence provenance.




\end{document}